Highlights

## FoodMem: Near Real-time and Precise Food Video Segmentation

Ahmad AlMughrabi, Adrián Galán, Ricardo Marques, Petia Radeva

- Introduces FoodMem, the first near-real-time food video segmentation framework.

- Leverages a segmentation transformer and memory model for mask refinement.

- Outperforms FoodSAM, the state-of-the-art, across diverse conditions with 2.5% higher mAP.

- Achieves 58x faster processing with minimal hardware resources.

- Provides a novel annotated food dataset with challenging scenarios.

# FoodMem: Near Real-time and Precise Food Video Segmentation

Ahmad **AlMughrabi**[a],  Adrián **Galán**[a],  Ricardo **Marques**[b] and  Petia **Radeva**[a,c]

[a]*Universitat de Barcelona, Gran Via de les Corts Catalanes, 585, Barcelona, 08007, Spain*
[b]*Interactive Technologies Group (GTI), Universitat Pompeu Fabra, Carrer de Tànger, 122-140, Barcelona, 08018, Spain*
[c]*Institut de Neurociències, University of Barcelona, Passeig de la Vall d'Hebron, 171, Barcelona, 08035, Spain*

## ABSTRACT

Food segmentation, including in videos, is vital for addressing real-world health, agriculture, and food biotechnology issues. Current limitations lead to inaccurate nutritional analysis, inefficient crop management, and suboptimal food processing, impacting food security and public health. Improving segmentation techniques can enhance dietary assessments, agricultural productivity, and the food production process. This study introduces the development of a robust framework for high-quality, near-real-time segmentation and tracking of food items in videos, using minimal hardware resources. We present FoodMem, a novel framework designed to segment food items from video sequences of 360-degree unbounded scenes. FoodMem can consistently generate masks of food portions in a video sequence, overcoming the limitations of existing semantic segmentation models, such as flickering and prohibitive inference speeds in video processing contexts. To address these issues, FoodMem leverages a two-phase solution: a transformer segmentation phase to create initial segmentation masks and a memory-based tracking phase to monitor food masks in complex scenes. Our framework outperforms current state-of-the-art food segmentation models, yielding superior performance across various conditions, such as camera angles, lighting, reflections, scene complexity, and food diversity[1]. This results in reduced segmentation noise, elimination of artifacts, and completion of missing segments. We also introduce a new annotated food dataset encompassing challenging scenarios absent in previous benchmarks. Extensive experiments conducted on MetaFood3D, Nutrition5k, and Vegetables & Fruits datasets demonstrate that FoodMem enhances the state-of-the-art by 2.5% mean average precision in food video segmentation and is 58× faster on average. The source code is available at:[2].

## 1. Introduction

Video Object Segmentation (VOS) [17] stands as a prevalent task within the domain of computer vision, finding applications in various fields, including object recognition, scene comprehension, medical imaging, and enhancing filter effects in video communication. While automated methodologies leveraging pretrained models for object segmentation are widely utilized, interactive user input is frequently incorporated to annotate novel training datasets or accomplish precise rotoscoping for intricate footage, particularly within visual effects contexts. This becomes particularly beneficial when videos present challenging lighting conditions and dynamic scenes or necessitate partial region segmentation. Although automatic VOS techniques strive to delineate entire objects with clearly defined semantic boundaries, Interactive Video Object Segmentation (IVOS) and Semi-supervised Video Object Segmentation (SSVOS) approaches [18, 24] offer enhanced flexibility. These typically entail the utilization of scribble or contour drawing interfaces for manual refinement.

Cutting-edge IVOS and SSVOS methodologies rely on memory-based models [1, 10] and have shown impressive segmentation outcomes in complex scenes through user-supplied mask annotations. However, these techniques are primarily designed to improve individual annotation performances [4, 5, 11, 22], making them unsuitable for practical production scenarios. They tend to over-segment recognized semantic contours (such as individuals, hair, faces, and entire objects) while struggling to accurately delineate partial regions (such as parts of a person's face or a dog's tail), and face challenges with lighting and extreme object orientations. As a result, when given only one annotated frame, inconsistent segmentations arise due to the inherent ambiguity in the object's appearance, particularly when subjected to significant variations in viewing angles and complex lighting conditions. This limitation limits the scalability of these techniques, as it is unclear which frame annotations are a priority, especially in extended sequences. A new SSVOS framework, XMem++ [1], stores annotated frames for reference and handles multiple frame annotations iteratively. XMem++ [1] adopted XMem [4] as a backbone. It improves object segmentation in complex scenes with fewer annotated frames without re-training. Using attention-based frame suggestion, it predicts the next annotations, supporting both sparse and dense annotations for better quality scaling. The system achieves real-time video segmentation with instant consideration of frame annotations.

---

[1]More details in the supplementary material.
[2]https://amughrabi.github.io/foodmem

✉ ahmad.almughrabi@ub.edu (A. AlMughrabi); ricardo.marques@upf.edu (R. Marques); petia.ivanova@ub.edu (P. Radeva)
ORCID(s): 0000-0002-9336-3200 (A. AlMughrabi); 0000-0001-8261-4409 (R. Marques); 0000-0003-0047-5172 (P. Radeva)

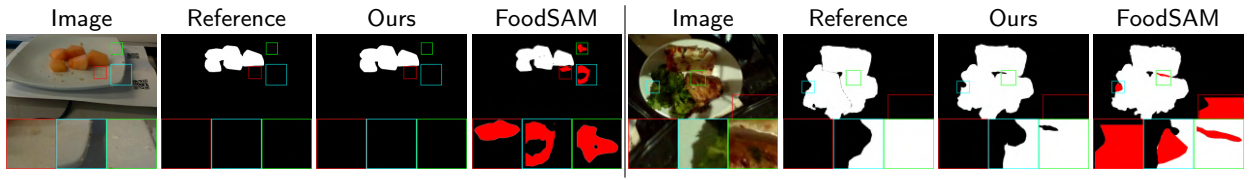

**Figure 1:** Comparison among original images, ground truth masks, masks produced by FoodMem, and masks produced by FoodSAM using the Nutrition5k dataset. The color red region highlights any artifacts or missing segments. Our method demonstrates robust performance under complex food geometry and low-light conditions.

In contrast, a novel zero-shot framework, FoodSAM [9], is designed to integrate the original semantic mask with category-agnostic masks generated by the Segment Anything (SAM) [8]. While SAM exhibits impressive food image segmentation capabilities, its lack of class-specific information limits its effectiveness [9]. Conversely, conventional segmentation methods maintain category information but often sacrifice segmentation quality [9]. FoodSAM proposes fusing the original segmentation output with SAM's generated masks to enhance semantic segmentation quality. Identifying the mask's category based on its predominant elements represents a novel and practical approach to improving the semantic segmentation process [9]. Still, FoodSAM fails to generalize the segmentation across multiple frames in a given video; for instance, the estimated mask IDs (i.e., mask colors) are assigned differently for the same food portion in different frames for the same scene. Moreover, FoodSAM fails to segment perfectly and generates masks from different camera views, such as missing food parts, and segments non-food objects, such as plates. Furthermore, FoodSAM is slow, which makes it an unsuitable scene segmentation framework for production scenarios.

To overcome these limitations, we propose FoodMem, a hybrid framework that uniquely leverages the semantic segmentation capabilities of SeTR with the memory-based tracking efficiency of XMem++. This fusion allows FoodMem to track and refine food object segmentations across video frames with minimal annotations while maintaining robustness under complex lighting and food geometries with varying viewing angles. Our framework generates one or a few annotated images as a first step and then tracks them throughout the scene. We further introduce a dataset for benchmarking purposes, which includes the new challenging scenes and food use cases, including: 1. **different capturing settings** such as 360º food scenes; high and low speed capturing; blurry images; complex and simple backgrounds; different illumination; 2. **food diversity and complexity** such as basic ingredients (e.g., apple) or complex ingredients (e.g., cuisine). 3. **bounded and unbounded scenes** such as scene has been captured by Mobile phone in a free movement; while some scene has been captured by Intel Realsense camera in a restricted movement. We extensively evaluate and compare our framework on our dataset, which is built from real-world and challenging datasets [19, 20] that are publicly available to show how our framework outperforms the current state-of-the-art.

## 2. Related Work

Various approaches have been proposed to tackle the food segmentation challenge. FoodSAM [9] has achieved a groundbreaking milestone as the first system to accomplish instance segmentation, panoptic segmentation, and promptable segmentation. FoodSAM utilizes the ViTh [6] as SAM's image encoder, employing the same hyperparameters outlined in the original paper. Object detection is performed using UniDet [25] with Unified learned COIM RS200. For semantic segmentation, SeTR [23] serves as FoodSAM baseline, incorporating ViT-16/B as the encoder and MLA as a decoder in FoodSeg103 [21], which SeTR focuses on extracting richer feature representations. Following a comprehensive evaluation of the FoodSeg103 [21] and UECFoodPix Complete [12] benchmarks, FoodSAM has outperformed the state-of-the-art methods on both datasets, demonstrating the exceptional potential of SAM as a powerful tool for food image segmentation. However, FoodSAM [9] exhibits considerable performance drawbacks (significant latency and high memory consumption), attributable to the integration of multiple models within its framework. Also, FoodSAM cannot track food items throughout a video, as it segments each image independently.

In contrast, the k-Means++ [16] technique is extensively utilized in various food segmentation tasks [20, 16] due to its ease of implementation, high speed, and the fact that it is an analytical solution requiring no training, prior knowledge or GPU to run. This method is particularly effective when the camera is positioned above the food, and the table background is simple, containing only food, a fork, a spoon, and a plate. It primarily serves as an RGB image segmentation method employing K-means clustering to differentiate the object from the background, although additional processing, such as median filtering, is necessary to address issues like uneven illumination. Subsequent steps involve converting the clustered image to grayscale, applying median filtering, Otsu thresholding, and morphological operations (erosion, dilation, opening, and closing) to enhance segmentation quality and accuracy in volume prediction, particularly for brightly colored food objects.

For video segmentation, DEVA [3] is a decoupled video segmentation approach that uses task-specific image-level segmentation and class/task-agnostic bi-directional temporal propagation, allowing it to 'track anything' without needing video data training for each individual task. This method only requires training an image-level model for the target task and a universal temporal propagation model once, which generalizes across tasks. DEVA outperforms end-to-end approaches in data-scarce tasks, including large-vocabulary video panoptic segmentation, open-world video segmentation, referring video segmentation, and unsupervised video object segmentation. Using DEVA with the prompt "food" demonstrates good overall model performance and speed; however, it exhibits limitations in recognizing a diverse range of food items and consumes a high amount of memory.

To overcome these limitations, we propose FoodMem, a production-grade video food segmentation framework that accepts a video or a set of images and segments food in near-real-time performance and memory-friendly. In contrast to prior approaches, which process each frame independently, FoodMem utilizes stored annotations to ensure consistent tracking of food items across video sequences. The combination facilitates near-real-time segmentation performance and enhances memory efficiency, thereby improving the overall robustness and applicability of food segmentation in diverse production environments. Our main contributions are as follows: 1. We build a novel near-real-time food segmentation architecture for videos that combine Segmentation Transformer (SeTR) [23] and memory-based model [1] frameworks as a first exploration for video food segmentation. 2. We introduce a novel dataset tailored for food image segmentation tasks. Our dataset comprises a comprehensive selection of dishes sourced from the Nutrition5k [20] dataset, encompassing 31 distinct dishes with a total of 1356 annotated frames. Additionally, we include 11 dishes from the Vegetable and Fruits [16] dataset, augmented with 2308 annotated frames. Our dataset features 42 diverse dishes, accompanied by 3664 meticulously annotated frames. We believe this expansive dataset is a valuable resource for advancing research in video food segmentation and related computer vision tasks. 3. We conducted an extensive series of experiments on our dataset to assess the effectiveness and flexibility of our framework against the baselines. 4. Our framework outperforms the state-of-the-art performance in video food segmentation for 2.5% mean average precision with similar recall. 5. Our framework is 58 times faster than the baseline's [9] inference time.

The rest of the paper is structured as follows: We present the theoretical background in Sec. 3. We define our proposed methodology in Sec. 3.2. A thorough set of experimental results is presented in Sec. 4. Finally, we present our conclusions and future work in Sec. 5.

## 3. Proposed Methodology

This section details the structure of our proposed method for automated food portion segmentation and tracking. Sec. 3.1 provides a high-level overview of the methodology, with reference to Fig. 2. Initially, we employ the SeTR model [23] to generate one or several food masks per scene. Subsequently, these masks are tracked within complex scenes using the XMem++ model [1]. The following sections delve into the specifics of each step in our approach.

### 3.1. Overview

Our framework leverages a two-phase framework, combining SeTR and memory-based frameworks to achieve efficient and rapid video segmentation, addressing the critical time complexity issue. This combination is inherently complex due to their distinct architectural designs and operational methodologies. Merging these models is non-trivial and involves significant challenges in aligning their functionalities and optimizing their interoperability. We have developed robust solutions to seamlessly integrate these components, ensuring precise segmentation and tracking while maintaining computational efficiency.

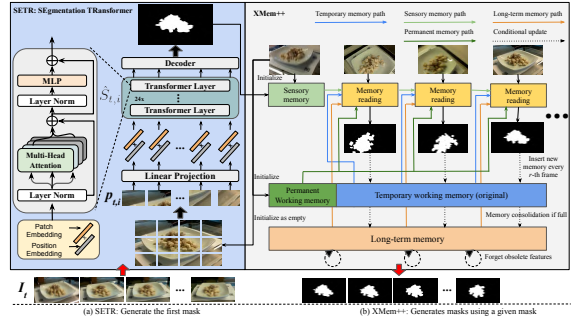

**Figure 2:** FoodMem model architecture. We used a single image input for simplicity. Our two-stage framework (a) shows the SeTR framework, where it accepts an image and generates a mask, followed by (b) a memory-based framework, which accepts the mask and a set of images as a given input and produces masks for all frames.

### 3.2. Our Proposal: FoodMem

Our methodology starts with semantic segmentation with the pre-trained SeTR model for food segmentation in video data. Specifically, we utilized the Sequence-to-Sequence Transformer (SeTR) model [23], leveraging pre-trained weights to perform initial semantic segmentation on the first $n$ frames $\{I_t\}_{t=1}^n$ of the video sequence. Here, $I_t$ denotes the $t$-th frame of the video sequence. The video frames are partitioned into $N$ non-overlapping patches $\{p_{t,i}\}_{i=1}^N$, each of size $P \times P \times C$. In this context, $P$ represents the height and width of each square patch, and $C$ denotes the number of color channels (e.g., $C = 3$ for RGB images). These patches are embedded into feature vectors $E_{t,i} \in \mathbb{R}^D$ using a linear projection, where $D$ denotes the dimensionality of the embedded feature vector $E_{t,i} = W_p \cdot \text{flatten}(p_{t,i}) + b_p$, where $W_p$ and $b_p$ are the learnable parameters of the linear projection. These feature vectors are augmented with positional embeddings $P_i$ to encode spatial relationships. The sequence of embedded patches $Z_{t,0} = \{E_{t,i} + P_i\}_{i=1}^N$ undergoes multiple layers of transformer encoding, leveraging pre-trained weights to enhance feature extraction capabilities. Here, $Z_{t,0}$ represents the initial sequence of embedded patches with positional embeddings. The output of the transformer encoder $Z_{t,L}$ is subsequently fed into the transformer decoder to refine feature representations. Segmentation predictions are generated by applying a linear projection followed by a softmax activation to produce the probability distribution [1, 4]: $\hat{S}_{t,i} = \text{softmax}(W_s Z_{t,L}^i + b_s)$, for each patch $i$. In this case, $Z_{t,L}$, where $L$ denotes the number of layers in the transformer encoder, is the output of the transformer encoder after $L$ layers, and $W_s$ and $b_s$ are the learnable parameters for the linear projection in the decoder. The predicted segmentation probability distribution for patch $i$ in frame $t$ is denoted by $\hat{S}_{t,i}$, as shown in Fig. 2 (a).

We leverage Long-Term Tracking with XMem++ [1, 4] to maintain temporal coherence in segmentation across frames. We integrate the XMem model without training, focusing on memory mechanisms for tracking and refining segmentation masks. For each frame $I_t$, feature maps $F_t = \phi(I_t)$ are extracted and used to populate Short-Term Memory (STM) and Long-Term Memory (LTM). Here, $\phi$ denotes the feature extraction function, and $F_t$ denotes the feature map extracted from frame $I_t$. STM retains recent feature maps [1, 4]: $\text{STM}_t = \{F_{t-n}, \ldots, F_t\}$, while LTM accumulates feature maps selectively based on their significance [1, 4]: $\text{LTM}_{t+1} = \text{LTM}_t \cup \{F_t\}$, To track segmentation masks over time, similarities $S_{t,i}$ between the current feature map $F_t$ and memory features $M_i$ are computed using dot products, normalized to obtain attention weights [1, 4]: $\alpha_{t,i} = \frac{\exp(S_{t,i})}{\sum_j \exp(S_{t,j})}$, where $S_{t,i}$ represents the similarity between the current feature map $F_t$ and memory feature $M_i$, and $\alpha_{t,i}$ denotes the attention weight for memory feature $M_i$. The memory read $R_t$ is computed as a weighted sum of memory features [23]: $R_t = \sum_i \alpha_{t,i} M_i$. Fig. 2 (b) illustrates Long-Term Tracking method, which takes the mask and a set of images as input and generates masks for all frames.

### 3.3. FoodMask: Our proposed dataset

We utilized the Nutrition5k [20] dataset, which comprises approximately 5000 dishes, offering a wide array of food items for analysis. For comprehensive testing, we carefully selected 31 plates from this dataset. Each plate was chosen to encompass diverse compositions of ingredients, ensuring a broad scope for evaluating our segmentation model's performance across various food categories. Each selected plate contains between 130 and 560 frames, totaling several hundred per plate. To streamline processing and reduce computational overhead, we employed an image processing pipeline to minimize the number of frames, specifically Imagededup [7]. This reduction consolidated the frames to approximately 20 to 65 keyframes per plate, preserving critical visual information essential for accurate segmentation analysis. Additionally, we incorporated the Vegetables & Fruits (V&F) dataset into our evaluation [16]. Like Nutrition5k, V&F is a substantial dataset containing multiple dishes and frames. We strategically selected a subset of dishes from both datasets based on food variation, video complexity, and ingredient diversity to ensure efficient validation of our framework. Similarly, we employed a near-image similarity approach [7], eliminating highly overlapped images and optimizing dataset integrity. We manually annotated each selected dish frame using LabelMe [15]. This meticulous annotation resulted in a dataset comprising 31 dishes from Nutrition5k with 1356 annotated frames and 11 dishes from V&F with 2308 annotated frames, totaling 42 dishes with 3664 annotated frames. This curated dataset of masks from Nutrition5K and V & F we call *FoodMask*, forms a robust foundation for rigorous testing and validation of our segmentation framework, facilitating accurate assessment across diverse culinary compositions.

The MetaFood3D (MTF3D) dataset [2] includes twenty food scenes across three difficulty levels: easy (8 scenes, 200 images), medium (7 scenes, 30 images), and hard (1 image per scene). Each scene contains food masks, depth images, a reference board, QR codes, and metadata for detailed evaluation [1].

Our methodology ensures robust food segmentation in video data by systematically integrating pre-trained SeTR for initial segmentation and memory-based XMem for tracking without retraining. Evaluated using mean average precision and recall, our approach provides quantitative measures of segmentation accuracy and temporal coherence, demonstrating its efficacy in real-world applications such as the FoodSeg103 dataset. This combination of transformer-based segmentation and memory-based tracking advances state-of-the-art video segmentation techniques, offering reliable food recognition and analysis tools[1].

## 4. Experimental Results

Our framework is evaluated on two public datasets: Nutrition5k dataset [20] and Vegetables and Fruits dataset [16]. We apply two quality metrics: mAP and recall metrics.

### 4.1. Implementation settings

We ran the experiments using an NVIDIA GeForce RTX 2080 Ti with 11GB of VRAM. For SeTR, we set one-shot image segmentation; for keyframe selection, we set the hamming distance to 12[1].

### 4.2. Quality Metrics

We outline the quality metrics used to evaluate the performance and effectiveness of our framework. These metrics are crucial for assessing the model's accuracy, efficiency, and reliability. Mean Average Precision (mAP) [14] and recall [13] are the primary metrics for evaluating the FoodMem model's performance. By focusing on these metrics, we aim to ensure that our model is accurate and reliable for practical applications in food segmentation tasks. These metrics provide insights into the model's capability to detect and segment food items effectively, which is essential for dietary assessment and culinary automation applications[1].

### 4.3. FoodMem Results

We extensively evaluated our model using Nutrition 5k dataset, Vegetables & Fruits, and MetaFood3D datasets. For the experiments on the datasets, we compared it to FoodSAM [9]. We reproduced the DEVA, kMean++, and FoodSAM results using the same quality metrics. Table 1 presents mAP and recall for each dataset. Briefly, we calculated quality metrics for each image in each considered scene to ensure consistency with previous works. We then determine the quality metrics at the scene level by averaging the quality of all images of the same scene. Next, we average the quality metrics across all scenes to compute the final quality values per method. This process is repeated for the two datasets. Notably, our model requires only one day to complete training, which is 58 times faster than the baseline, as shown in Fig. 3. The qualitative results on the Nutrition 5k are shown in Fig. 1. Additionally, Fig. 4 shows the qualitative results on the V&F dataset. The figures show that our model excels in texture details, artifact correction, and missing data handling across different scene parts, surpassing other models. Table 1 shows the quantitative results of our comparison. The table shows that our model performs better in food segmentation on different datasets[1].

### 4.4. Ablation study

To understand the factors contributing to FoodMem's success, we studied the impact of altering the masks generated by SeTR. Specifically, we examined the changes resulting from increasing the number of masks to 3, 6, or 9. We aimed

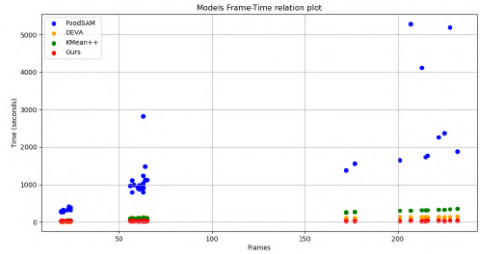

**Figure 3:** The inference time for the baselines and our framework. Our framework shows a stable inference time across a different number of frames. Our inference time is 58 times faster than the baseline.

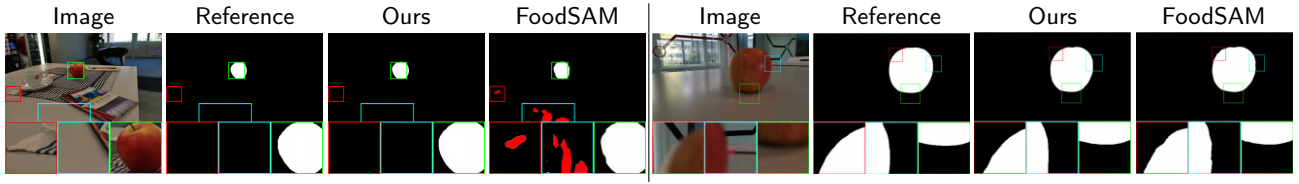

**Figure 4:** Comparison between original images, ground truth masks, masks generated by FoodMem, and masks generated by FoodSAM using Vegetables & Fruits dataset. Our method demonstrates robust performance under blurry (caused by fast speed capturing) and far-near conditions[1].

#### Table 1

Comparison of mAP and recall scores achieved by different models on two datasets: MetaFood3D, Nutrition5k, and V&F. The models evaluated include FoodSAM, DEVA, kMean++, and our framework. We also apply our method to each image individually for fairness*. All results are reproduced (best results in red, second-best in orange, and third-best in yellow) Time complexity, GPU memory usage (averaged over 67 images), and parameters for each evaluation model[1]. †This dataset forms part of the FoodMask collection, in which we manually annotate the ground truth masks.

| Methods | mAP↑ | | | Recall↑ | | | Model(s) Size | Complexity | |
|---|---|---|---|---|---|---|---|---|---|
| | N5k† | V&F† | MTF3D | N5k† | V&F† | MTF3D | | Speed↓ | Memory↓ |
| FoodSAM | 0.9192 | 0.8914 | 0.9348 | 0.7752 | 0.9441 | 0.9973 | 636M | 22m33s | 12.5G |
| DEVA | 0.8825 | 0.8548 | 0.2734 | 0.7301 | 0.9328 | 0.9511 | 241M | 25s | 9.0G |
| kMean++ | 0.4232 | 0.4361 | 0.5572 | 0.6467 | 0.9245 | 0.9718 | - | 15s | - |
| SeTR* | 0.9170 | 0.8306 | 0.8946 | 0.9788 | 0.9906 | 0.9524 | 723M | 14m51s | 6.25G |
| Ours | 0.9098 | 0.9499 | 0.9453 | 0.7708 | 0.9469 | 0.9941 | 785M | 25s | 6.25G |

#### Table 2

Comparison of recall and mAP scores obtained by various FoodMem configurations on two datasets, Nutrition5k, MetaFood3D, and V&F. The configurations include settings with 1 mask, 3 masks, 6 masks, and 9 masks. All results are reproduced (best results in red, second-best in orange, and third-best in yellow). †This dataset forms part of the FoodMask collection, in which we manually annotate the ground truth masks.

| Masks | mAP↑ | | | Recall↑ | | |
|---|---|---|---|---|---|---|
| | N5k† | V&F† | MTF3D | N5k† | V&F† | MTF3D |
| 1 mask | 0.9098 | 0.9499 | 0.9453 | 0.7708 | 0.9469 | 0.9941 |
| 3 masks | 0.9025 | 0.9027 | 0.9781 | 0.7688 | 0.9419 | 0.9953 |
| 6 masks | 0.9005 | 0.9124 | 0.9282 | 0.7663 | 0.9438 | 0.9923 |
| 9 masks | 0.9082 | 0.9050 | 0.9213 | 0.7690 | 0.9430 | 0.9916 |

to present both qualitative and quantitative outcomes of these modifications. For qualitative analysis, Fig. 5 compared the masks visually. For quantitative analysis, Table 2 assessed the execution time and evaluated performance using mean average precision and recall metrics.

### 4.5. Discussion

While our framework demonstrates robust performance in maintaining temporal coherence in segmentation tasks, our findings indicate that it generates artifacts in the resulting masks when multiple clean masks are provided. This issue arises from the model's reliance on feature similarity for memory read operations. When the features of multiple clean masks are highly similar, the dot product-based similarity measure used by XMem++ can lead to ambiguous attention weight assignments, resulting in incorrect weighting of memory features. Consequently, portions of the mask may be incorrectly assigned or blended, reducing segmentation accuracy and causing visual inconsistencies, as shown in Fig. 4 for Pear and Apple objects[1].

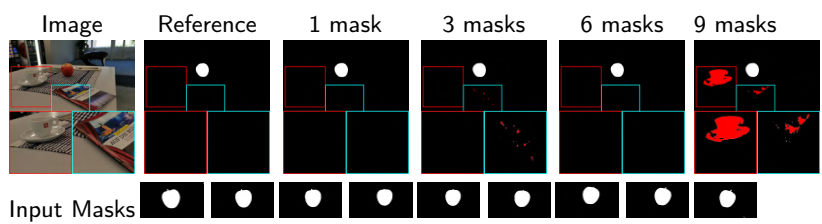

**Figure 5:** Comparison of original images, ground truth masks, and masks generated by SeTR with configurations of 1, 3, 6, and 9 masks using Vegetables & Fruits dataset. We show the input masks for each scene.

These artifacts are particularly problematic in complex scenes where precise segmentation is critical. To mitigate these limitations, further research is needed to enhance the feature differentiation capabilities of XMem++. Potential approaches include improving the feature extraction process to generate more distinct feature vectors, incorporating advanced similarity measures to better handle subtle differences between features, or refining the memory update mechanism to selectively store and retrieve more discriminative features. Addressing this limitation is crucial for improving the reliability and accuracy of segmentation in complex video sequences.

*Limitations.* Similar to the baselines, several fundamental limitations of our framework are identified, which are pivotal for comprehending our findings and guiding future research directions. 1. Leveraging XMem++ introduces inherited constraints related to its algorithmic design and computational requirements, which could impact scalability and efficiency when processing large datasets or high-resolution (e.g., 4k) videos. 2. Our framework exhibits sensitivity to lighting conditions, particularly in low-light environments or scenarios with significant shadows. This issue is most pronounced in the first frames processed by SeTR, where it can pose challenges in accurately detecting and segmenting food objects before the memory-based mechanism stabilizes the segmentation across subsequent frames. 3. Our framework can generate artifacts when processing multiple clean masks with similar features, resulting in ambiguous attention weights and decreased accuracy in scenes with visually similar food items, as shown in our Ablation study. Addressing these limitations will enhance the framework's reliability and applicability across varied real-world conditions and datasets in future research endeavors[1].

## 5. Conclusions and Future Work

Our proposed framework represents a significant advancement in video semantic segmentation of food items. We have introduced a novel architecture that effectively leverages the strengths of SeTR and memory-based models, leading to state-of-the-art performance in food video segmentation. Additionally, we have contributed a comprehensive dataset tailored for this task, providing a valuable resource for future research. Our experimental results demonstrate the framework's superior accuracy and efficiency, making it well-suited for practical food recognition and analysis applications. In future work, we aim to explore advanced preprocessing and feature extraction techniques and feature differentiation strategies to mitigate our limitations and improve the robustness of our framework in such scenarios.

## CRediT authorship contribution statement

**Ahmad AlMughrabi:** This work was partially funded by the EU project MUSAE (No. 01070421), 2021-SGR-01094 (AGAUR), Icrea Academia'2022 (Generalitat de Catalunya), Robo STEAM (2022-1-BG01-KA220-VET000089434, Erasmus+ EU), DeepSense (ACE053/22/000029, ACCIÓ), and Grants PID2022141566NB-I00 (IDEATE), PDC2022-133642-I00 (DeepFoodVol), and CNS2022-135480 (A-BMC) funded by MICIU/AEI/10.13039/501100011033, by FEDER (UE), and by European Union NextGenerationEU/ PRTR. A. AlMughrabi acknowledges the support of FPI Becas, MICINN, Spain.

# FoodMem: Near Real-time and Precise Food Video Segmentation

Ahmad **AlMughrabi**[a], Adrián **Galán**[a], Ricardo **Marques**[a,b] and Petia **Radeva**[a,c]

[a]*Universitat de Barcelona, Gran Via de les Corts Catalanes, 585, Barcelona, 08007, Spain*

[b]*Computer Vision Center, Generalitat de Catalunya and the Universitat Autònoma de Barcelona, Campus UAB, Edifici O, s/n, Barcelona, 08193, Spain*

[c]*Institut de Neurociències, University of Barcelona, Passeig de la Vall d'Hebron, 171, Barcelona, 08035, Spain*

## 1. Proposed Methodology

In this section, Algo. 1 meticulously outlines the steps of our method, offering a detailed breakdown of each phase. Our process starts with semantic segmentation using the SeTR model on the video sequence's first $n$ frames. Each frame is divided into patches, transformed into feature vectors, and then processed through transformer encoding to make segmentation predictions. We leverage long-term tracking, which uses Short-Term and Long-Term Memory mechanisms to maintain consistent segmentation across frames over time. Our approach enables us to refine segmentation masks accurately and coherently throughout the video sequence, as described in the algorithm.

---

**Algorithm 1** Our framework pseudo-code for our proposed methodology.

---

1: **Input:** Video sequence $\{I_t\}_{t=1}^{T}$, number of initial frames $n$
2: **Output:** Segmentation masks for each frame $\{\hat{S}_t\}_{t=1}^{T}$
3: **Initialization:**
4: **for** $t = 1$ to $n$ **do**
5:     Partition $I_t$ into patches $\{p_{t,i}\}_{i=1}^{N}$ of size $P \times P \times C$
6:     **for** each patch $p_{t,i}$ **do**
7:         Embed patch into feature vector $E_{t,i} = W_p \cdot \text{flatten}(p_{t,i}) + b_p$
8:         Augment feature vector with positional embedding $P_i$
9:         Form initial sequence $Z_{t,0} = \{E_{t,i} + P_i\}_{i=1}^{N}$
10:     **end for**
11:     Apply transformer encoder with $L$ layers to $Z_{t,0}$
12:     Obtain encoder output $Z_{t,L}$
13:     Refine feature representations with transformer decoder
14:     **for** each patch $i$ **do**
15:         Generate segmentation prediction $\hat{S}_{t,i} = \text{softmax}(W_s Z_{t,L}^{i} + b_s)$
16:     **end for**
17: **end for**
18: **Long-Term Tracking:**
19: **for** $t = n + 1$ to $T$ **do**
20:     Extract feature map $F_t = \phi(I_t)$
21:     Update Short-Term Memory (STM): $STM_t = \{F_{t-n}, \ldots, F_t\}$
22:     Update Long-Term Memory (LTM): $LTM_{t+1} = LTM_t \cup \{F_t\}$
23:     **for** each memory feature $M_i$ in $STM_t$ and $LTM_t$ **do**
24:         Compute similarity $S_{t,i}$ between $F_t$ and $M_i$
25:         Normalize to obtain attention weight $\alpha_{t,i} = \frac{\exp(S_{t,i})}{\sum_j \exp(S_{t,j})}$
26:     **end for**
27:     Compute memory read $R_t = \sum_i \alpha_{t,i} M_i$
28:     Generate segmentation mask for $I_t$ using $R_t$
29: **end for**
30: **Return:** Segmentation masks $\{\hat{S}_t\}_{t=1}^{T}$

---

## 1.1. Leveraging Models Discussion

Leveraging SeTR and XMem++ in our framework is more than just combining architectures; it dramatically improves both the speed and precision of the segmentation process. By leveraging SeTR for accurate keyframe

---

✉ ahmad.almughrabi@ub.edu (A. AlMughrabi); ricardo.marques@ub.edu (R. Marques); petia.ivanova@ub.edu (P. Radeva)
ORCID(s): 0000-0002-9336-3200 (A. AlMughrabi); 0000-0001-8261-4409 (R. Marques); 0000-0003-0047-5172 (P. Radeva)

segmentation and XMem++ for memory-driven temporal processing, we ensure that our segmentation outputs are precise spatially and temporally consistent. This integration is also highly efficient (as shown in the qualitative and quantitative results), allowing our framework to achieve an average speedup of 58× over the state-of-the-art methods, making it ideal for near-real-time applications. Furthermore, the framework shows a 2.5% enhancement in mean average precision (mAP) compared to current techniques, indicating its capability to deliver more precise segmentation masks while processing frames significantly faster. This improvement is especially notable when dealing with lengthy video sequences with intricate scene dynamics, where maintaining temporal coherence and efficiency is vital.

## 2. Experimental Results

We present additional qualitative comparisons for the baseline methods. Fig. 1 and Fig. 2 compare our approach with FoodSAM on the V&F and Nutrition 5k datasets. Fig. 3 and Fig. 4 show comparisons with DEVA, while Fig 5 and Fig. 6 compare our method to kMean++ on the same datasets. The qualitative results, captured under free camera movement, reveal notable differences in segmentation performance across various methods. In Fig. 1 and Fig. 2, while FoodSAM performs reasonably well on the Nutrition 5k dataset, it introduces artifacts such as holes and missing regions, affecting the segmentation accuracy. In contrast, our method produces complete and artifact-free masks, effectively handling these challenges. Similarly, in Fig. 3 and Fig. 4, DEVA shows segmentation inaccuracies, including missing parts and artifacts, across the Nutrition 5k and Vegetables and Fruits datasets, whereas our approach consistently generates more accurate segmentations under similar conditions. Lastly, Fig. 5 and Fig. 6 demonstrate that kMean++, when tested under complex food geometries and challenging lighting, struggles with finer details, leading to segmentation errors. Moreover, we add additional qualitative experiments using the MetaFood3D dataset as shown in Fig. 8. Our method, however, remains robust, providing consistent and high-quality results despite the unconstrained camera movements and varying conditions. These findings collectively highlight the superiority of our approach in handling segmentation tasks under free camera motion across diverse datasets. Fig. 16 shows qualitative comparisons on motion blur using the Vegetables and Fruits dataset.

For further insights on SeTR training can be found in Table 4, which presents qualitative results leveraging FoodSeg103 alongside details such as model size. Meanwhile, Fig. 4 illustrates the qualitative outcomes derived from the same dataset. The main metrics used to assess the Segmenter's performance are mIoU (mean Intersection over Union for each class), mACC (mean accuracy for all classes), and aAcc (accuracy on all pixels) [10].

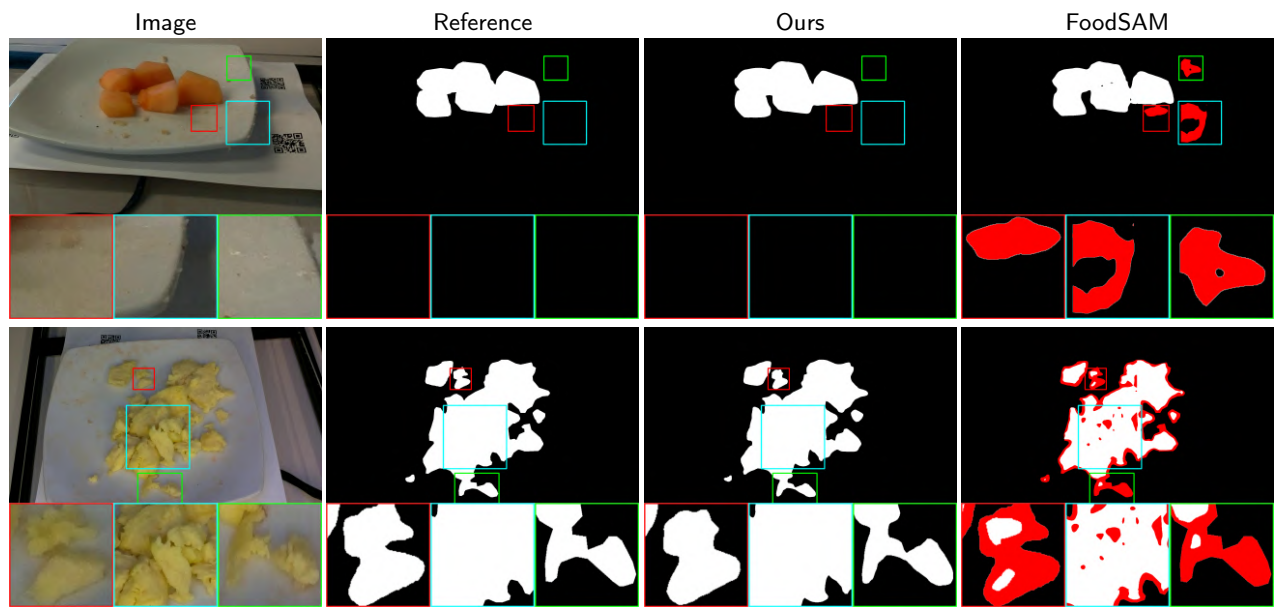

**Figure 1:** Comparison between original images, ground truth masks, masks generated by FoodMem, and masks generated by FoodSAM. The dataset used is Nutrition5k. The color red region highlights any artifacts or missing segments.

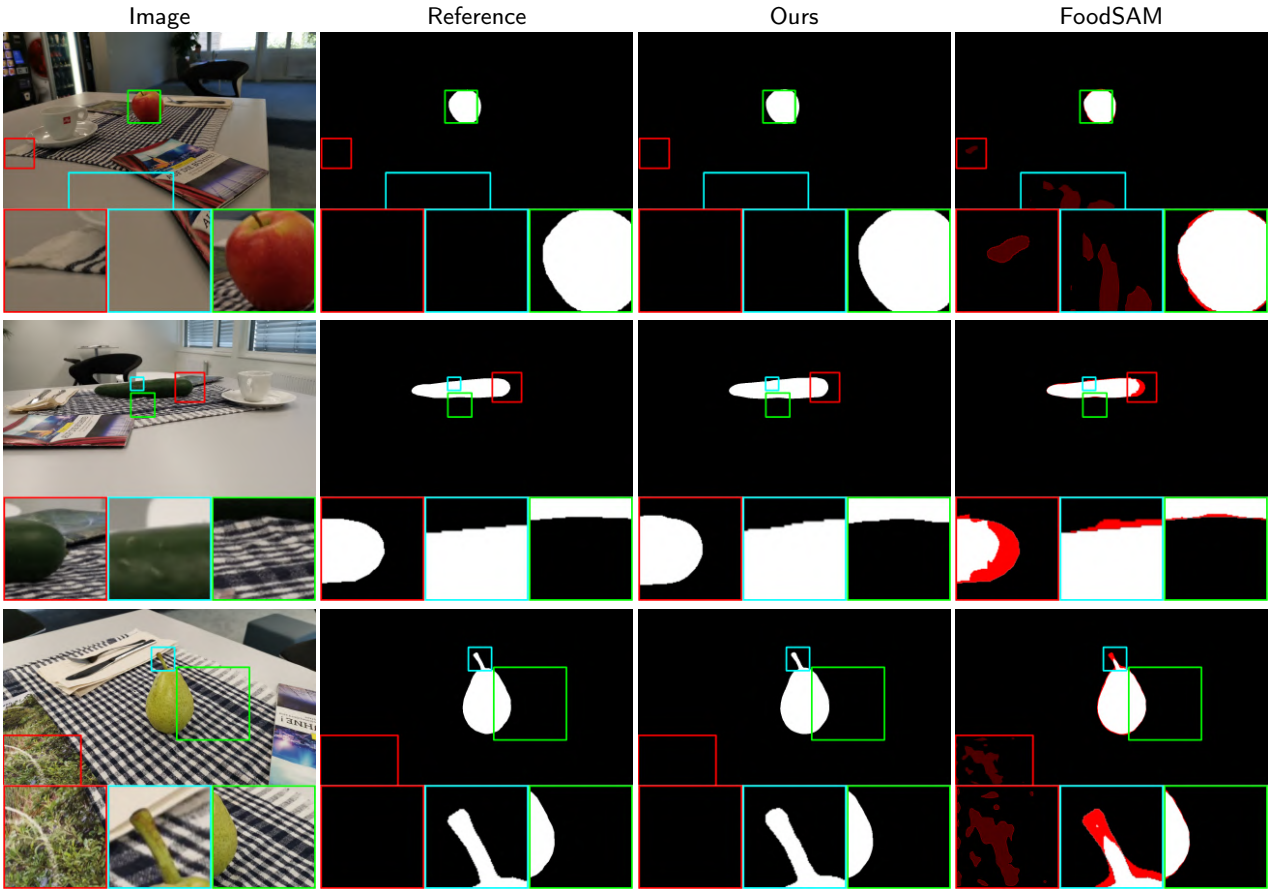

**Figure 2:** Comparison between original images, ground truth masks, masks generated by FoodMem, and masks generated by FoodSAM. The dataset used is Vegetables & Fruits. The color red region highlights any artifacts or missing segments.

In order to further exemplify the efficacy of our approach, we present a qualitative comparison illustrated in Fig. 9. While Table 1. (main paper) delineates the average performance across multiple scenes, this figure concentrates on a singular scene (ID = 1) from the MetaFood3D dataset, thereby providing a comprehensive visual analysis. It is noteworthy that for this particular scene, SeTR* achieves a mAP of 0.7843, whereas FoodMem attains a considerably superior mAP of 0.9398 (i.e., SeTR* with memory-based models). This comparative analysis underscores the enhanced accuracy of FoodMem when employed in conjunction with the tracking methodology, thereby reinforcing the quantitative findings articulated in the main paper.

## 2.1. Implementation Settings

We conducted additional experiments to show the memory usage for DEVA, FoodSAM, and ours on a given food scene containing 67 images from Nutrition 5k; each image dimension has 1920x1080 width and height. FoodMem shows minimal memory usage with time; FoodSAM can only process 4 images with higher memory usage. Our method uses 2.13x less memory than FoodSAM, as demonstrated in Fig. 10. The SeTR model, which builds upon ViT-16/B, extracts feature maps from the 12th transformer encoder, followed by two sets of convolutional layers for prediction. The ViT-16/B model [4], a transformer-based architecture pre-trained on the ImageNet-21k dataset. ViT-16/B consists of 12 transformer encoders, each equipped with 12-head self-attention mechanisms. For positional embeddings, we apply bilinear interpolation to reinitialize the pre-trained embeddings. All other segmentor components follow their default configurations and are initialized randomly.

For the SeTR learning parameters, we use the default parameters as [10]. Each image shall be resized to a fixed dimension of $2049 \times 1024$ pixels while adhering to a ratio range between 0.5 and 2.0. Subsequently, a patch measuring $768 \times 768$ pixels is cropped from these resized images, applying random horizontal flipping and color jitter. The models were trained over 80,000 iterations utilizing 8 images per batch, optimized through Stochastic Gradient Descent (SGD) solvers with a momentum coefficient set at 0.9 and a weight decay of 0.0005. The initial learning rate for the SeTR is established at 1e-3. In accordance with established settings [6, 9, 10], the learning rate decreases by a factor of 0.9 according to a polynomial decay schedule. For the sake of simplicity, hard negative mining is not employed during the training process, and SeTR operates on the widely adopted MMSegmentation platform [3, 10]. Hamming distance is utilized in the context of near image similarity using the PHash algorithm, as implemented in the Imagededup framework by [7]. This metric measures the similarity between the current image and its potential matches in memory. The framework is tuning the Hamming distance threshold as 12 by default. For the learning parameters, DEVA utilizes the Segment Anything Model (SAM) with the ViT-B configuration (91M parameters) alongside Grounding DINO (172M parameters) and the DEVA model itself, which has 69.16M parameters. SAM provides three pre-trained image encoder configurations tailored to different scales: ViT-B (91M parameters), ViT-L (308M parameters), and ViT-H (636M parameters) [1, 5]. In our framework, SeTR, which has 723M parameters [10], employs the SAM ViT-B/16 configuration for image encoding, while XMem++, with 62M parameters, handles long-term temporal consistency. Additionally, FoodSAM leverages the SAM ViT-H configuration (636M parameters) [8]. These configurations demonstrate the varying parameter sizes and computational capacities of comparative methodologies.

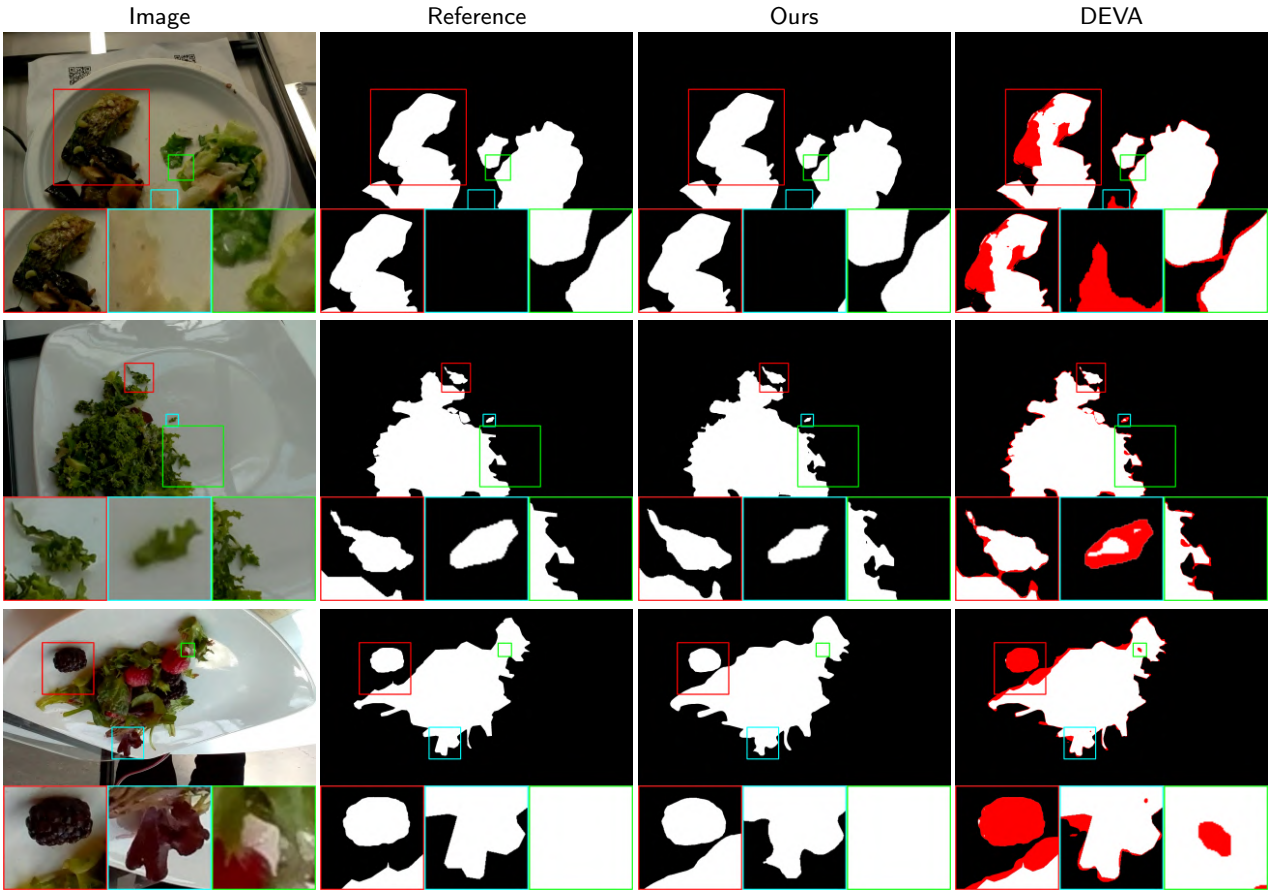

**Figure 3:** Comparison between original images, ground truth masks, masks generated by FoodMem, and masks generated by DEVA. The dataset used is Nutrition5k. The color red region highlights any artifacts or missing segments.

### 2.2. Datasets

The FoodSeg103 [10] dataset comprises 103 ingredient categories grouped into 15 overarching supercategories. We randomly assigned 70% (4,983 images) of the dataset for training and used the remaining 30% for testing. Detailed summary statistics for the training and testing sets are in Table 2. Our experiments involve training and evaluating with the FoodSeg103 dataset, which is used for in-domain training and evaluation. Table. 3 shows the total count of ingredients across all superclasses FoodSeg103.

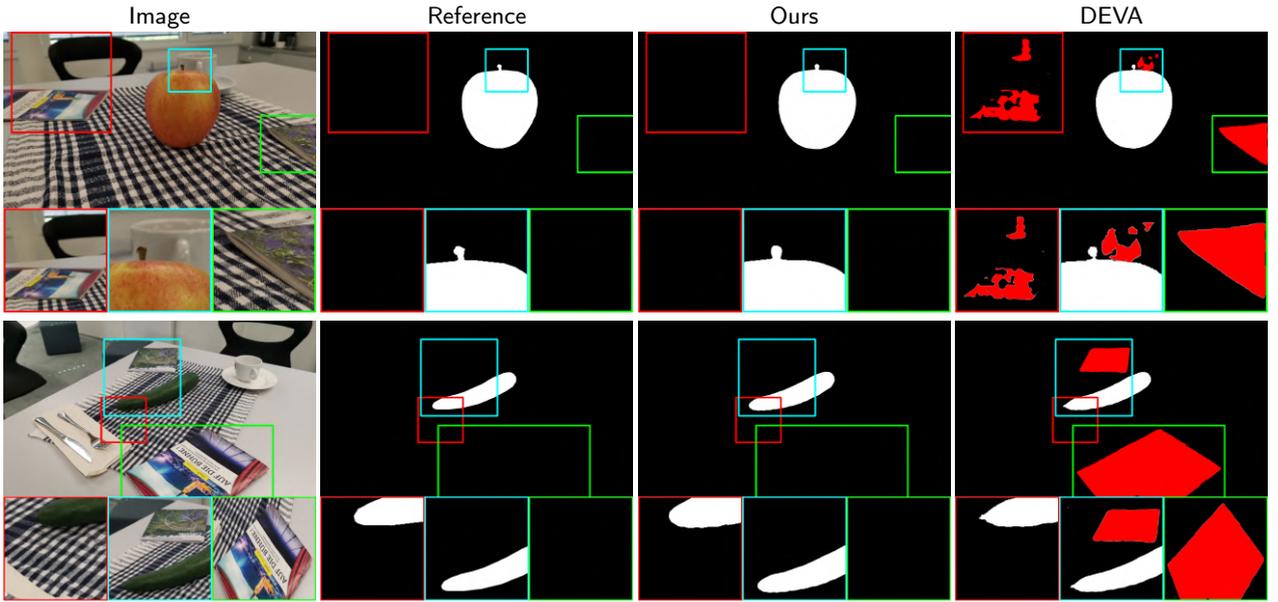

**Figure 4:** Comparison between original images, ground truth masks, masks generated by FoodMem, and masks generated by DEVA. The dataset used is Vegetables & Fruits. The color red region highlights any artifacts or missing segments.

**Table 1**
MetaFood3D dataset details and number of images for each scene.

| L | ID | Food Name | # Images |
|---|----|-----------|----------|
| E | 1 | Strawberry | 199 |
| | 2 | Cinnamon bun | 200 |
| | 3 | Pork rib | 200 |
| | 4 | Corn | 200 |
| | 5 | French toast | 200 |
| | 6 | Sandwich | 200 |
| | 7 | Burger | 200 |
| | 8 | Cake | 200 |
| | 9 | Blueberry muffin | 30 |
| M | 10 | Banana | 30 |
| | 11 | Salmon | 30 |
| | 12 | Steak | 30 |
| | 13 | Burrito | 30 |
| | 14 | Hotdog | 30 |
| | 15 | Chicken nugget | 30 |
| H | 16 | Everything bagel | 1 |
| | 17 | Croissant | 1 |
| | 18 | Shrimp | 1 |
| | 19 | Waffle | 1 |
| | 20 | Pizza | 1 |
| | | | 1620 |

The MetaFood3D [2] dataset comprises twenty distinct food scenes categorized into three levels of difficulty: easy, medium, and hard. The easy category encompasses eight scenes, each containing approximately 200 images. The medium category consists of seven scenes, with approximately 30 images each, whereas the hard category features only a single image per scene, as shown in Table. 1. Each scene includes food masks, depth images, a reference board (such as a chessboard), and QR codes. Additionally, a metadata file supplies detailed descriptions of each food object, facilitating a comprehensive evaluation across the varying difficulty levels, as shown in Fig. 11. The dataset is publicly accessible.

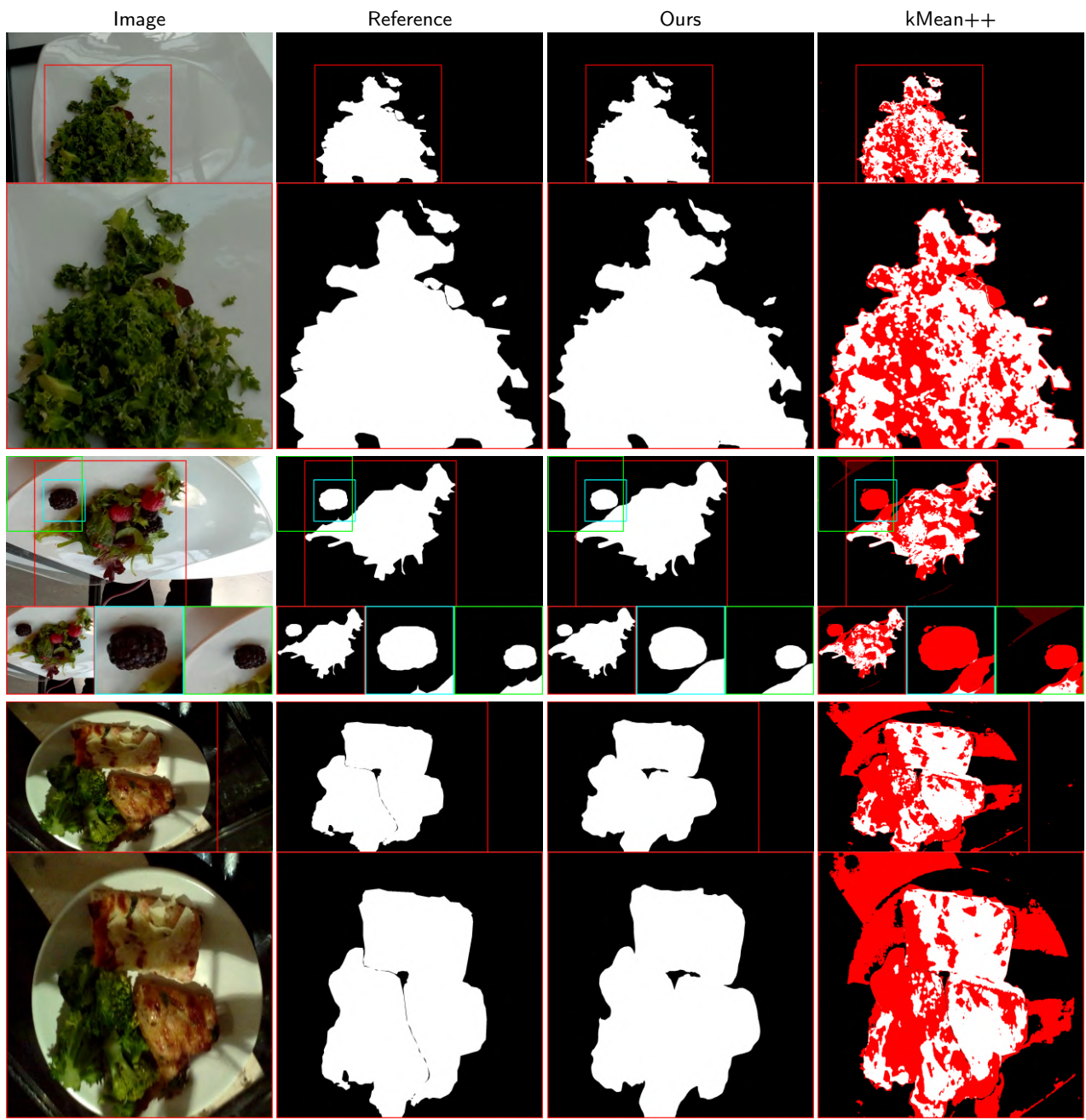

**Figure 5:** Comparison between original images, ground truth masks, masks generated by FoodMem, and masks generated by kMean++. The dataset used is Nutrition5k. The color red region highlights any artifacts or missing segments.

### 2.2.1. Datasets Characteristics

Our framework surpasses the current state-of-the-art models for food segmentation, exhibiting exceptional performance across diverse conditions. The Vegetables and Fruits and Metafood3D datasets, distinguished by free-motion capturing with unrestricted camera placements, underscore our model's capability to address unbounded scenes, which present significant challenges due to the variability in camera angles and perspectives as demonstrated in Fig. 12, Fig. 17, Fig. 18, and Fig. 19. Furthermore, these datasets highlight the robustness of our framework in effectively managing reflections from glossy surfaces, unpredictable lighting variations, and the complexity of motion blur, as shown in Fig. 17 and Fig. 18. In contrast, the Nutrition5k dataset, characterized by bounded scenes captured using fixed cameras and a custom scanning rig, exemplifies our model's adaptability to controlled environments. Additionally, it illustrates our capacity to accommodate a variety of lighting conditions (natural, artificial, and mixed) and diverse food items while maintaining effectiveness across multiple camera angles (e.g., overhead and side views). Collectively, these attributes demonstrate the resilience and versatility of our model in real-world applications.

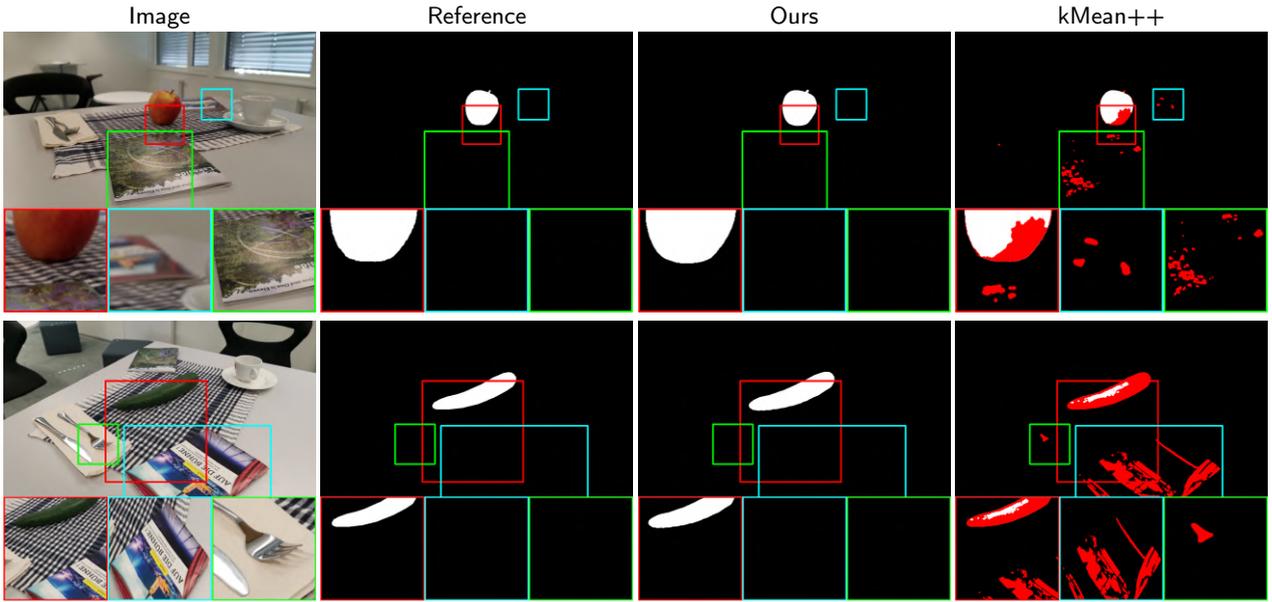

**Figure 6:** Comparison between original images, ground truth masks, masks generated by FoodMem, and masks generated by kMean++. The dataset used is Vegetables & Fruits. The color red region highlights any artifacts or missing segments.

**Table 2**
Training and testing set statistics for FoodSeg103.

| Dataset | #Images | | | #Ingredients | | |
|---|---|---|---|---|---|---|
| | Train | Test | Total | Train | Test | Total |
| FoodSeg103 | 4,983 | 2,135 | 7,118 | 29,530 | 12,567 | 42,097 |

**Table 3**
The total count of ingredients across all superclasses FoodSeg103.

| Superclasse | Number | Superclasse | Number | Superclasses | Number |
|---|---|---|---|---|---|
| Dessert | 3913 | Meat | 4956 | Soy | 148 |
| Beverage | 844 | Condiment | 1543 | Vegetable | 15719 |
| Nut | 912 | Seafood | 920 | Fungus | 592 |
| Egg | 424 | Soup | 121 | Salad | 23 |
| Fruit | 6007 | Main | 5634 | Others | 341 |

**Table 4**
Shows the quantitative results of the FoogSeg103 dataset.

| Method | mIoU | mAcc | Model Size |
|---|---|---|---|
| SeTR | 41.3 | 52.7 | 723M |

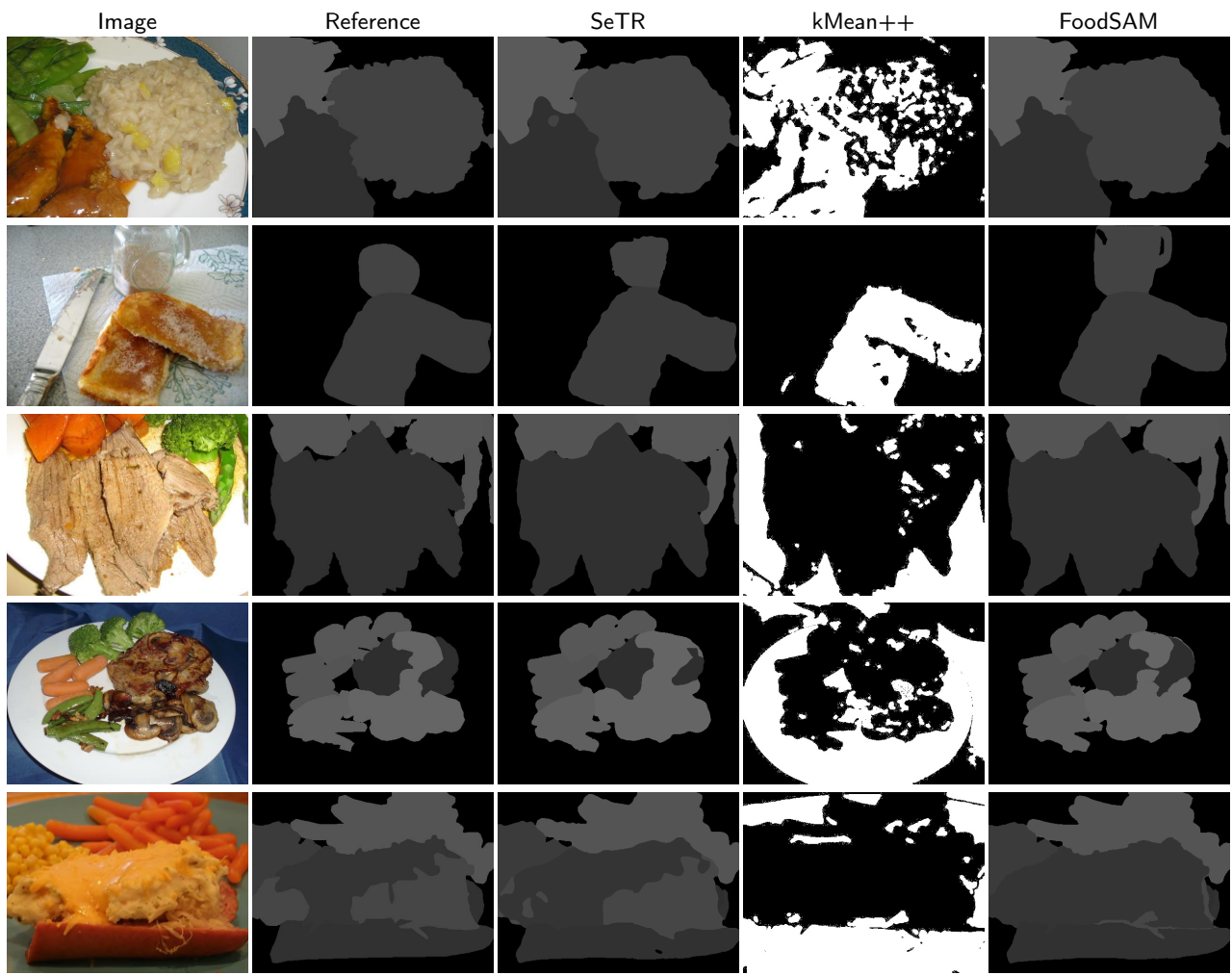

**Figure 7:** Comparison between original images, ground truth masks, masks generated by SeTR, kMean++, and FoodSAM using the FoodSeg103 dataset.

### 2.2.2. Dataset Utilization

Our framework's SeTR component underwent pre-training on the FoodSeg103 dataset, followed by testing on its set-to-set baseline performance. To confirm our framework generalizability, we expanded our evaluation to three other datasets: Vegetables & Fruits, Nutrition5k (Masks are created manually as an integral component of our FoodMask dataset for both datasets), and MetaFood3D. These datasets presented various challenges, such as irregular shapes, reflections, different lighting conditions, and unbounded scenes captured with free- and fixed-motion cameras. This thorough evaluation highlighted our framework's robustness and adaptability in real-world situations.

### 2.3. Ablation Study

In this section, we show additional ablation study. Fig. 13, Fig. 15, Fig. 20 show the ablation study when running our method using different input views for initialization (i.e, SeTR) before entering these masks for memory-tracking model (i.e., XMem++). We also present comparisons of the execution times for the various configurations of our framework. Highlighting the time taken by each configuration to process video frames, particularly in real-time applications. Table 5 displays the average execution time for each modified setting when processing the videos in our dataset.

### 2.4. Quality Metrics

We outline the quality metrics used to evaluate the performance and effectiveness of our framework.

**Table 5**
Average execution times of our framework's different settings. The settings include 1 mask, 3 masks, 6 masks and 9 masks. The inference time were recorded in the format of hours:minutes:seconds.

| Dataset | Frames range | 1 mask | 3 masks | 6 masks | 9 masks |
|---------|--------------|--------|---------|---------|---------|
| **Nutrition5k** | 19-65 | **00:00:25** | 00:00:35 | 00:00:50 | 00:01:05 |
| **V&F** | 172-232 | **00:00:31** | 00:00:39 | 00:00:45 | 00:00:56 |
| **MTF3D** | 30-200 | **00:00:27** | 00:00:33 | 00:00:47 | 00:01:03 |

### 2.4.1. Mean Average Precision

mAP is a comprehensive measure that combines precision and recall to evaluate the accuracy of object detection models. It calculates the average precision across different recall levels, providing a metric that reflects the model's performance in detecting food items. The mAP is defined as:

$$\text{mAP} = \frac{1}{N} \sum_{i=1}^{N} \text{AP}_i \tag{1}$$

where $N$ is the number of classes, and $\text{AP}_i$ is the average precision for class $i$. Using mAP, we can assess how well our model identifies and localizes food items in images, ensuring high accuracy and effectiveness.

### 2.4.2. Recall

measures the model's ability to correctly identify all relevant instances within a dataset. It is the ratio of true positive detections to the total number of relevant instances. Recall is defined as:

$$\text{Recall} = \frac{TP}{TP + FN} \tag{2}$$

where $TP$ is the number of true positives, and $FN$ is the number of false negatives. High recall indicates that the model effectively detects the majority of food items, minimizing the occurrence of missed detections. This metric is crucial for applications where it is essential to capture all instances of food items, such as in dietary assessment, where missing an item could lead to inaccurate dietary intake records.

### 2.4.3. Mean Intersection over Union

The **mean Intersection over Union (mIoU)** is a metric used to evaluate the overlap between predicted and ground truth segmentation masks. For each class $c$, IoU is calculated as:

$$\text{IoU}_c = \frac{|P_c \cap G_c|}{|P_c \cup G_c|} \tag{3}$$

where $P_c$ and $G_c$ are the predicted and ground truth pixels for class $c$, respectively. The mIoU is the average IoU across all classes:

$$\text{mIoU} = \frac{1}{C} \sum_{c=1}^{C} \text{IoU}_c \tag{4}$$

where $C$ is the total number of classes. mIoU measures segmentation accuracy by considering both false positives and false negatives.

### 2.4.4. Mean Accuracy

The **mean Accuracy (mAcc)** is the average accuracy across all classes, calculated by the proportion of correctly predicted pixels for each class. For class $c$, the accuracy is:

$$\text{Acc}_c = \frac{|P_c \cap G_c|}{|G_c|} \tag{5}$$

where $|P_c \cap G_c|$ is the number of correct predictions for class $c$, and $|G_c|$ is the total number of ground truth pixels for class $c$. The mAcc is the average of accuracies across all classes:

$$\text{mAcc} = \frac{1}{C} \sum_{c=1}^{C} \text{Acc}_c \tag{6}$$

mAcc evaluates the overall classification performance across classes, treating each class equally.

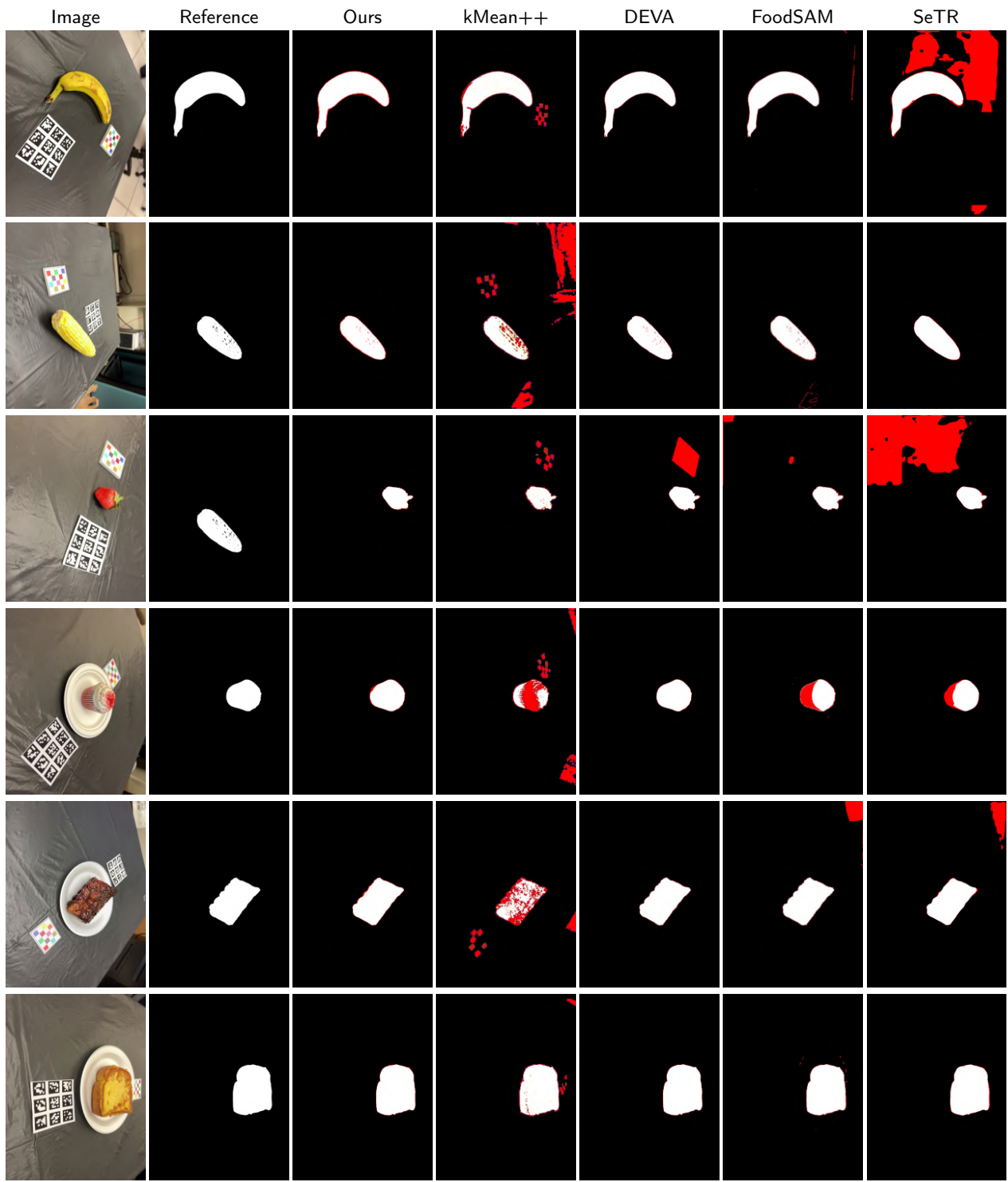

**Figure 8:** Comparison between original images, ground truth masks, masks generated by FoodMem, kMean++, DEVA, FoodSAM, and SeTR. The dataset used is MetaFood3D. The color red region highlights any artifacts or missing segments.

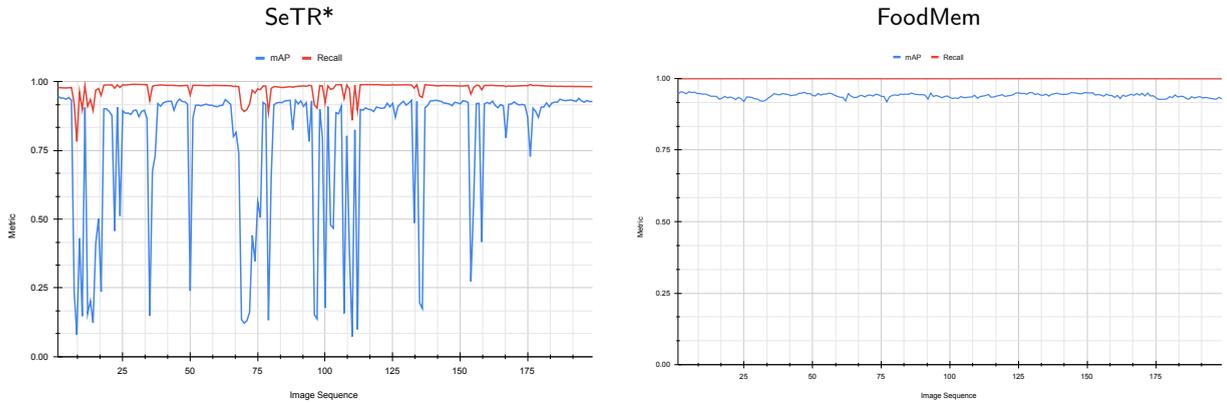

**Figure 9:** Comparison the provided text presents an analysis of the SeTR performance on individual frames. This illustration exemplifies the efficacy of our methodology when utilized in conjunction with the tracking technique. For this presentation, we exclusively employ the scene identified by ID = 1 from the MetaFood3D dataset. In contrast, Table 1 within the main paper provides the average results across a multitude of scenes. Notably, SeTR* achieves a mean Average Precision (mAP) of 0.7843 for this specific scene, whereas FoodMem reaches 0.9398.

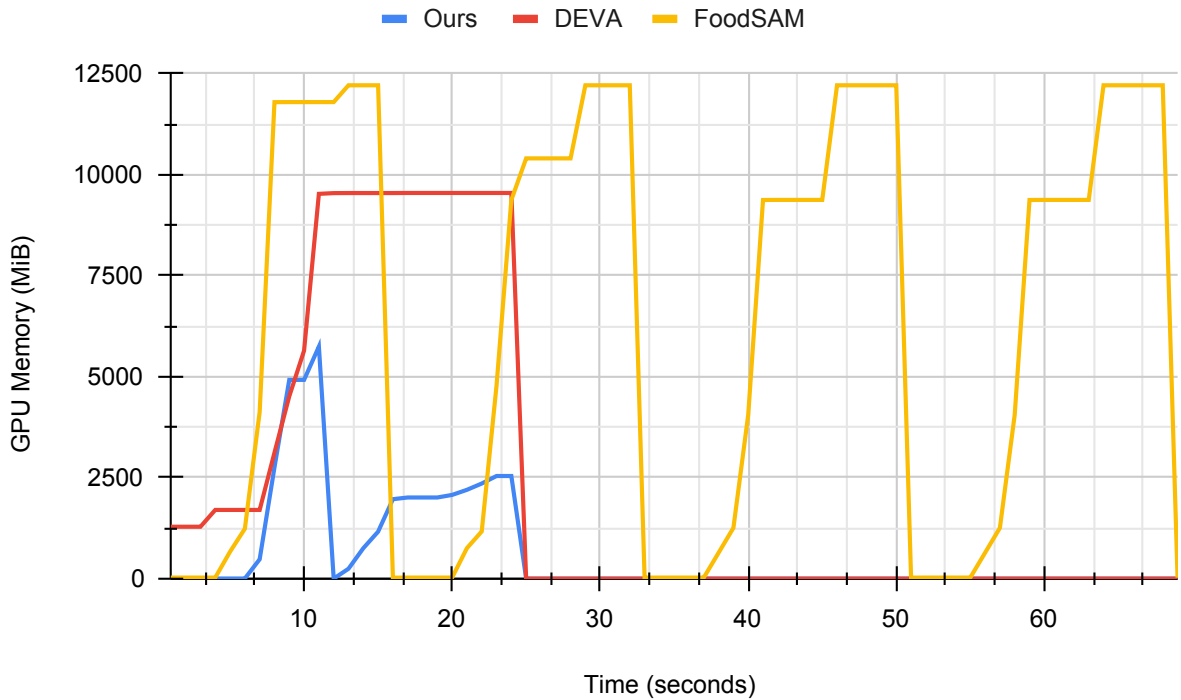

**Figure 10:** GPU memory usage on 67 images from the Nutrition 5k dataset (1920x1080 each). FoodMem has the lowest memory consumption, while FoodSAM processes only 4 images with higher usage. Our method reduces memory by 2.13X compared to FoodSAM.

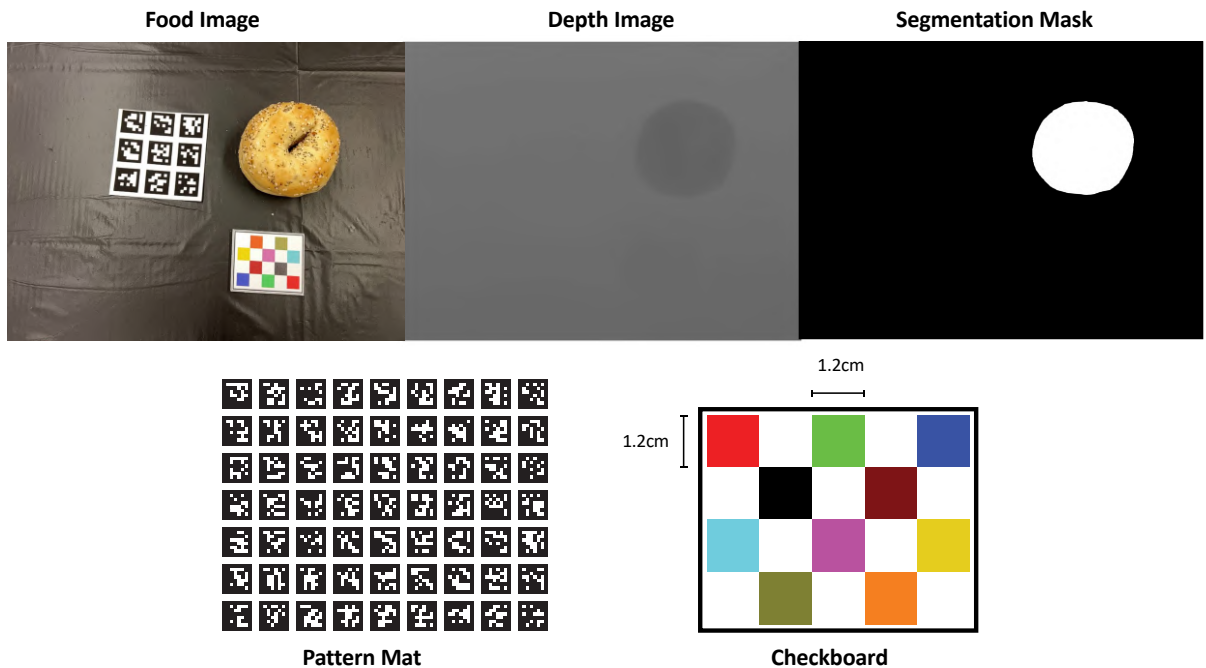

**Figure 11:** A sample image represents an RGB image, depth image, segmentation mask, pattern matrix, and checkboard.

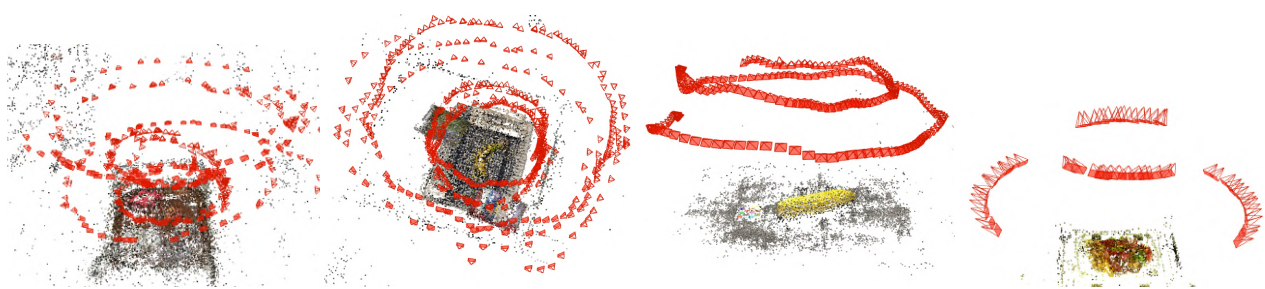

**Figure 12:** For illustration, camera locations, and orientations were estimated by Colmap for various bounded and unbounded scenes from the Vegetables and Fruits, Nutrition 5k, and MetaFood3D datasets. The initial two scenes on the left are derived from the Vegetables and Fruits dataset, the third scene originates from the MetaFood3D dataset, and the final scene is obtained from the Nutrition 5k dataset.

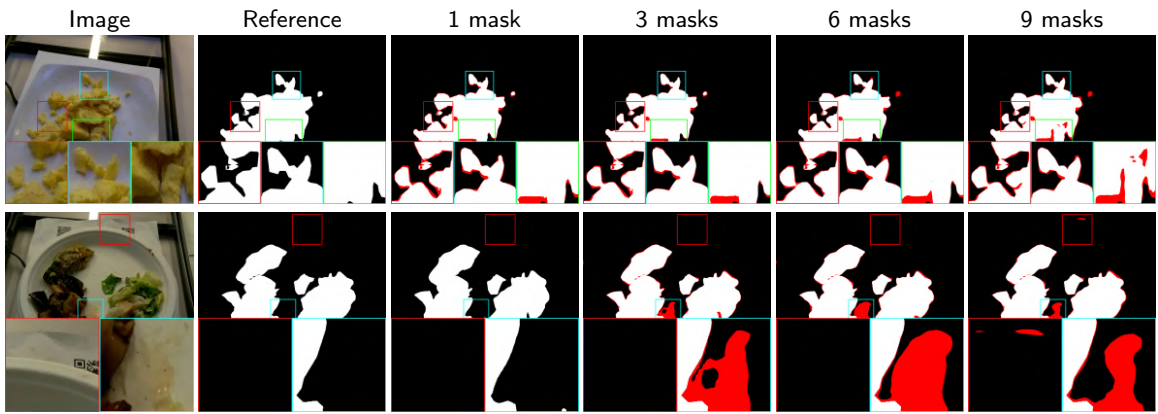

**Figure 13:** Comparison between original images, ground truth masks, 1 mask generated by SeTR, 3 masks generated by SeTR, 6 masks generated by SeTR, and 9 masks generated by SeTR. The dataset used is Nutrition5k.

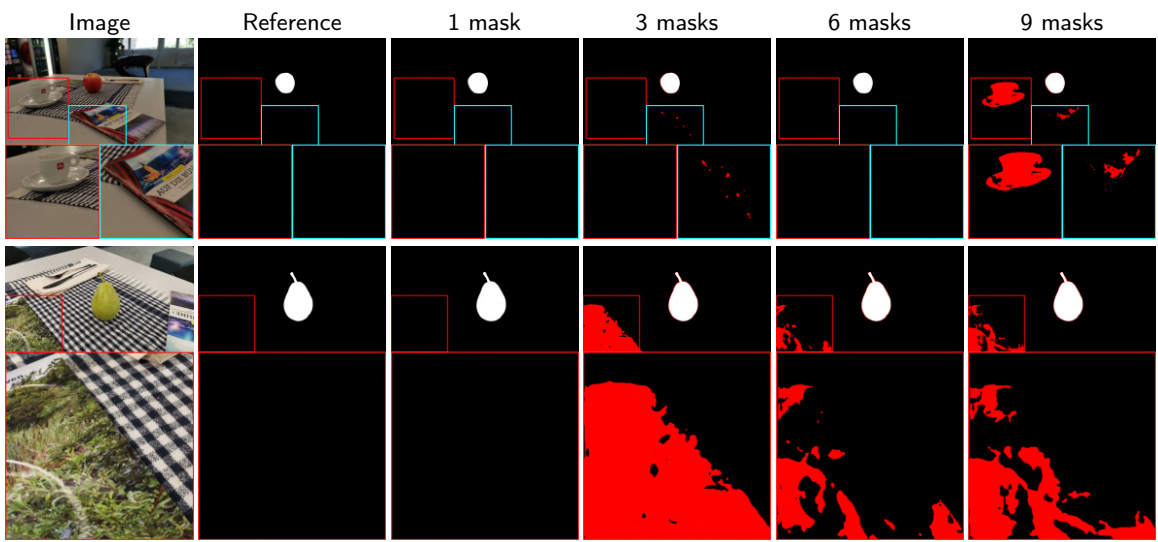

**Figure 14:** Comparison between original images, ground truth masks, 1 mask generated by SeTR, 3 masks generated by SeTR, 6 masks generated by SeTR, and 9 masks generated by SeTR. The dataset used is Vegetables & Fruits.

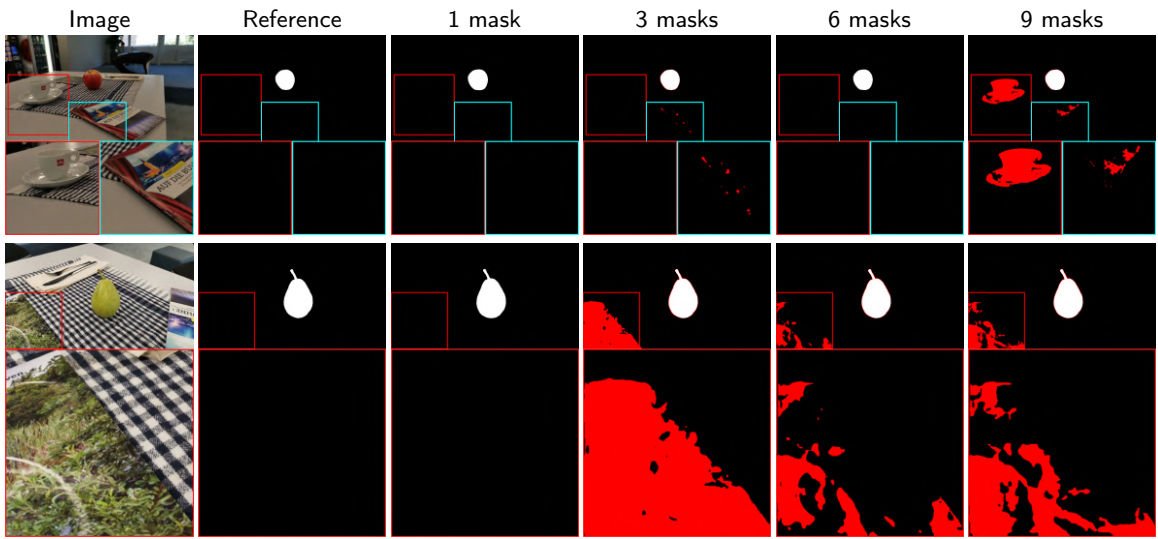

**Figure 15:** Comparison between original images, ground truth masks, 1 mask generated by SeTR, 3 masks generated by SeTR, 6 masks generated by SeTR, and 9 masks generated by SeTR. The dataset used is Vegetables & Fruits.

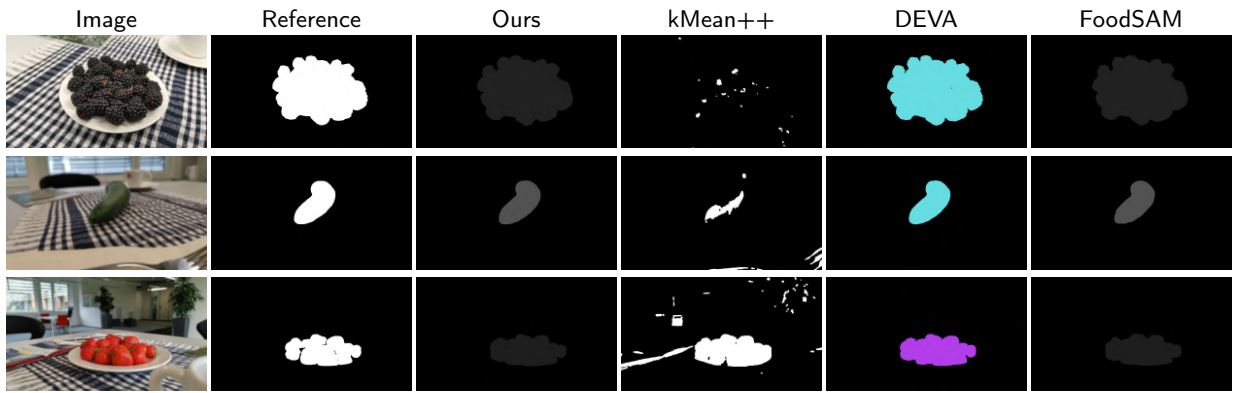

**Figure 16:** Comparison comparing original images, ground truth masks, and masks produced by FoodMem, kMean++, DEVA, and FoodSAM on blurred Vegetables and Fruits dataset images. The samples may appear sharper because the images have been resized to fit the article page.

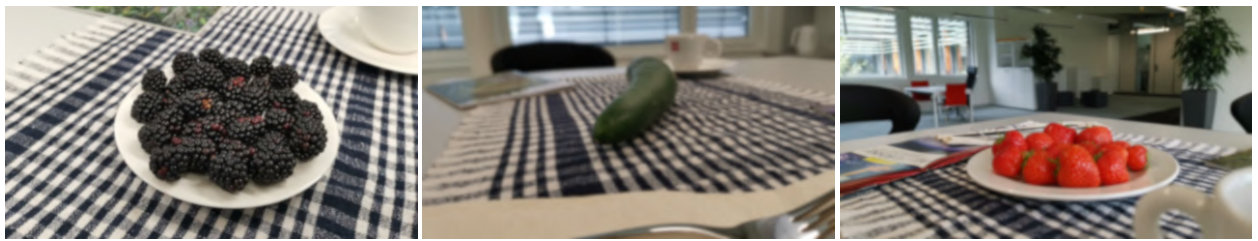

**Figure 17:** Examples we tackle the challenges of motion blur found in the Vegetables and Fruits datasets. This problem frequently occurs with cameras in free motion. The samples may appear sharper because the images have been resized to fit the article page.

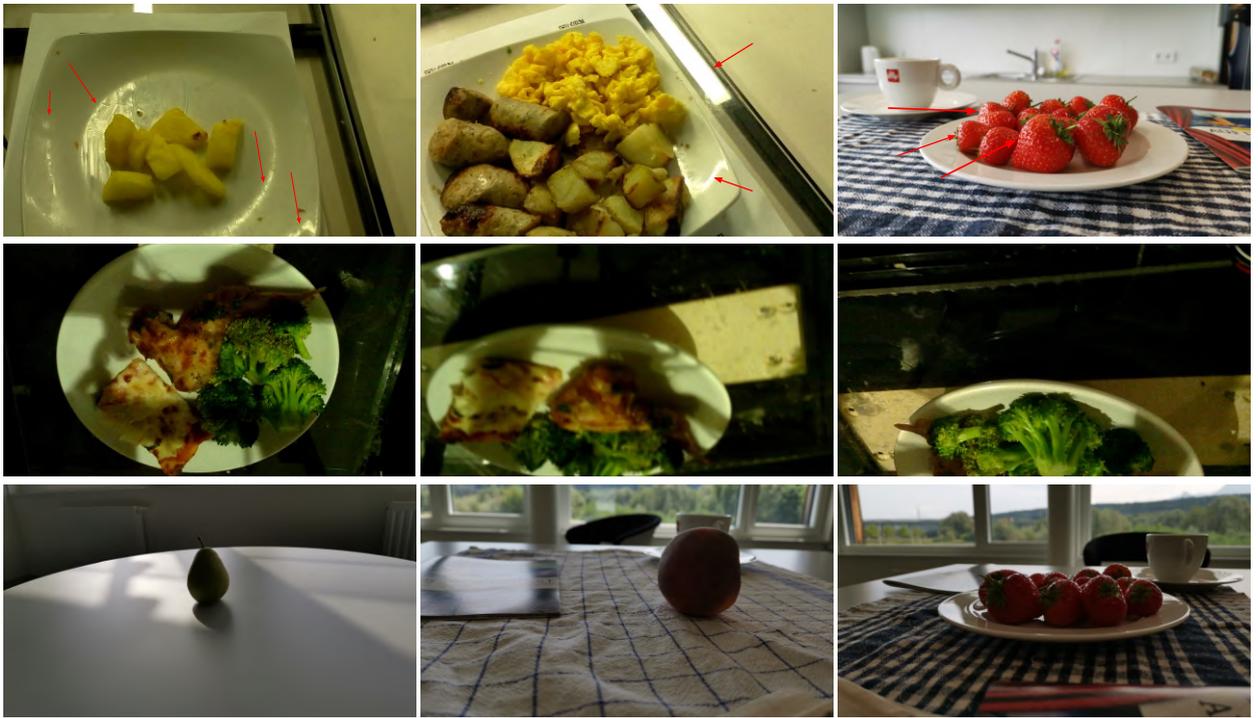

**Figure 18:** Examples we address the lighting challenges present in the Nutrition 5k and Vegetables and Fruits datasets. By utilizing both natural and artificial lighting, we emphasize the reflective surfaces. Additionally, we illustrate the low-light conditions that occur in some of these scenarios scenes.

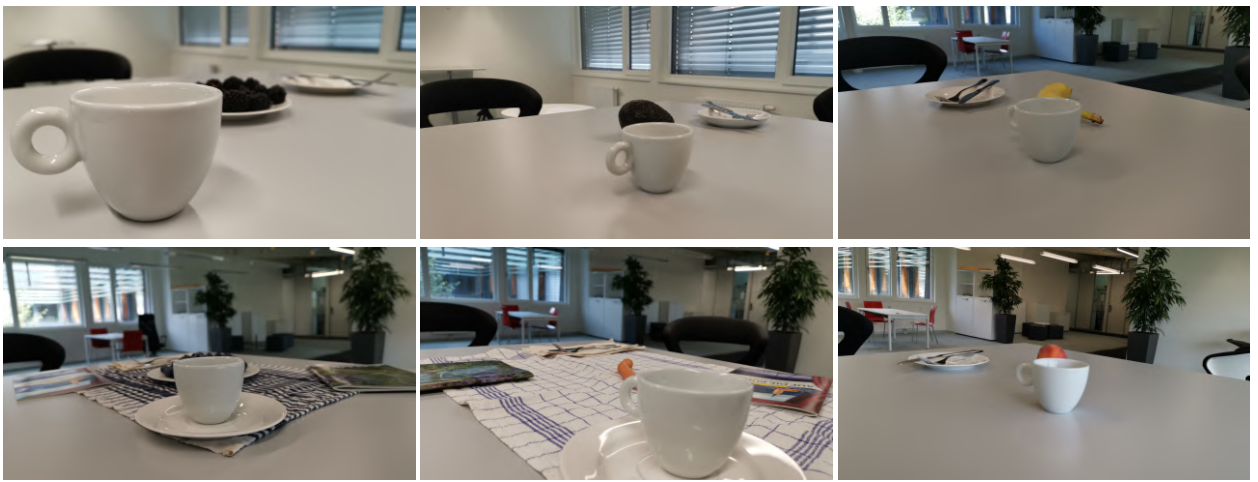

**Figure 19:** Examples of the occlusion issues present in the Vegetables and Fruits datasets are highlighted. The figure demonstrates that an obstacle object may obscure the food object depending on the camera position.

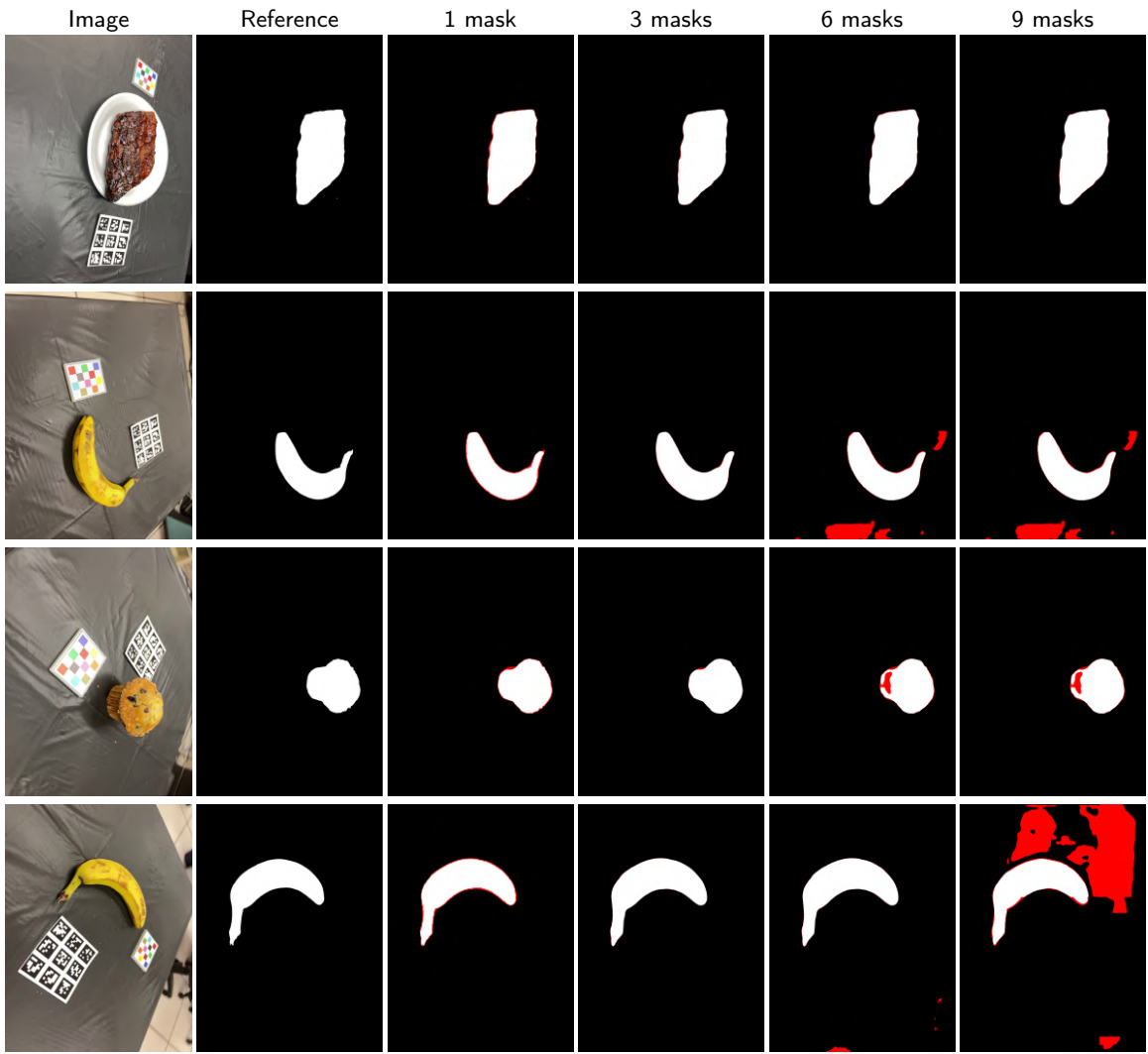

**Figure 20:** Comparison between original images, ground truth masks, 1 mask generated by SeTR, 3 masks generated by SeTR, 6 masks generated by SeTR, and 9 masks generated by SeTR. The dataset used is MetaFood3D.